# Privacy-Preserving Wavelet Neural Network with Fully Homomorphic Encryption

Syed Imtiaz Ahamed[1,2] and Vadlamani Ravi[1*]

[1]*Analytics Lab, Centre for AI and ML,*
*Institute for Development and Research in Banking Technology*
*Castle Hills, Masab Tank, Hyderabad 500057, India*
[2]*School of Computer and Information Sciences (SCIS), University of Hyderabad,*
*Hyderabad 500046, India*
syedahamed@idrbt.ac.in; vravi@idrbt.ac.in

**Abstract**

The main aim of Privacy-Preserving Machine Learning (PPML) is to protect the privacy and provide security to the data used in building Machine Learning models. There are various techniques in PPML such as Secure Multi-Party Computation, Differential Privacy, and Homomorphic Encryption (HE). The techniques are combined with various Machine Learning models and even Deep Learning Networks to protect the data privacy as well as the identity of the user. In this paper, we propose a fully homomorphic encrypted wavelet neural network to protect privacy and at the same time not compromise on the efficiency of the model. We tested the effectiveness of the proposed method on seven datasets taken from the finance and healthcare domains. The results show that our proposed model performs similarly to the unencrypted model.

## 1. Introduction

Machine Learning is being extensively used in almost every field such as healthcare, finance, education, intrusion detection, and even in recommendation systems (Al-Rubaie, & Chang, 2019). A lot of private data is stored in the databases and is openly utilized by the ML algorithms to build models from them. One of the major concerns in the application of ML models is the privacy and security of such private data. Organizations cannot simply ignore the privacy concerns of the data such as customers' Personal Identifiable Information (PII) and at the same time cannot stop analyzing such data because it would reap immense business and operational benefits to the organization.

On May 25, 2018, European Union (EU), brought into effect the toughest privacy and security law in the world called General Data Protection Regulation (GDPR) (Truong et.al., 2021). The law states that the organizations that violate the privacy and security standards will be imposed heavy fines of almost millions of euros. One more such law, namely, California Consumer Privacy Act (CCPA), allows the consumers in California the right to know about everything that a business collects about them, the right to delete the collected information, and the right to opt out of the sale of their information (Stallings, 2020). Similarly, Personal Data Protection Act (PDPA) enacted in Singapore protects personal Data (Chik, 2013).

---

[*] Corresponding author. Tel.: +914023294310; fax: +914023535157.

With such strict privacy laws, organizations are precluded from using private data freely. To overcome this problem, PPML provides different ways that will assure the customers that their data privacy will be protected and at the same time organizations can work on the private data and build better and more responsible ML Models.

There are different approaches in PPML and there is no single proven approach that is considered to be the best among all the approaches. For example, one of the approaches is Differential Privacy (DP) where the researchers can work on the peoples' personal information without disclosing their identity. But the drawback of DP is that it might lead to a loss in model accuracy. Similarly, another technique is called Secure Multi-Party Computation where multiple data owners can collaboratively train the model but this might result in high communication overhead or high computation overhead (Xu, Baracaldo, & Joshi, 2021).

One more approach is to secure the data using Homomorphic Encryption. It allows the computation to be performed on the encrypted data without the need for decryption. Partial Homomorphic Encryption (PHE), Somewhat Homomorphic Encryption (SWHE), and Fully Homomorphic Encryption (FHE) are the variations of Homomorphic Encryption. PHE allows an unlimited number of either additions or multiplications, SHE allows a limited number of arithmetic operations, and FHE allows an unlimited number of additions and multiplications on the encrypted data.

In this paper, we focus on the FHE which is considered to be the most secure technique compared to others. Here, we propose FHE based privacy-preserving Wavelet Neural Network (WNN). Thus we designed and implemented the secure WNN by ensuring that the data and all the trainable parameters in the network are fully homomorphic encrypted and also we get the results in an encrypted format.

The remaining part of the paper is structured as follows: In section 2, we discuss the related work regarding homomorphic encryption. Section 3 explains the proposed methodology and in section 4 the description of the datasets is presented. The results are discussed in Section 5 and finally, Section 6 concludes the paper. Appendix A consists of Tables presenting the features of datasets.

## 2 Literature Survey

Of late the idea of PPML, resulted in the enablement of privacy preservation in a few ML techniques. To start with privacy-preserving ridge regression was proposed in (Nikolaenko et.al., 2013), where the authors used a hybrid approach by combining a linear homomorphic encryption approach with Yao garbled circuits. Later, A fully homomorphic encrypted Convolution Neural Network was proposed in (Chabanne et al., 2017) and it was combined with the Cryptonets (Xie et.al., 2014) solution along with the batch normalization principles.

In Ref. (Chen, Gilad-Bachrach, & Han et al. 2018), the authors implemented fully homomorphic encryption Logistic Regression using the Fan-Vercauteren scheme implementation in the SEAL Library. In Ref. (Cheon, Kim, Kim, & Song, 2018). an ensemble gradient descent method was proposed for optimizing the coefficients in a homomorphically encrypted logistic regression, which resulted in the reduction of time complexity of the algorithm.

A secure Multi-layer perceptron was implemented in (Bellafqira, Coatrieux, Genin, & Cozic, 2019) which trains the homomorphically encrypted data on the cloud using the Paillier cryptosystem and makes use of two non-colluding severs. In Ref. (Nandakumar, Ratha, Pankanti, & Halevi, 2019), the authors trained a typical two-layered neural network on the encrypted data using fully homomorphic encryption with the help of an open-source library HElib (Halevi, & Shoup, 2020) for encryption.

In Ref. (Sun, Zhang, Liu, Yu, & Xie, 2020), the authors proposed an improved FHE scheme based on HElib and implemented a private hyper-plane decision-based classification and private Naïve Bayes Classification using the additive homomorphic and multiplicative homomorphic encryption. They implemented a private decision tree classification with the proposed FHE scheme.

Privacy-preserving Linear Regression model was implemented on distributed data in (Qiu, Gui & Zhao, 2020) which includes multiple clients and two non-colluding servers. The protocol consists of Paillier Homomorphic Encryption and data masking technique. In Ref. (Bonte, & Vercauteren, 2018), Privacy-Preserving Logistic Regression was implemented where the authors worked on somewhat homomorphic encryption based on the scheme of Fan and Vercauteren (Fan, & Vercauteren, 2012), and the model is trained on the encrypted data.

## 3. Proposed Methodology

In this section, the concepts of homomorphic encryption and its types along with CKKS scheme (Benaissa, Retiat, Cebere, & Belfedhal, 2021) which we employed for implementing the FHE are explained. Later, we explain the original unencrypted WNN and describe our proposed Privacy-Preserving WNN in detail along with a block diagram.

## 3.1 Homomorphic Encryption

Homomorphic Encryption is a special type of encryption scheme which allows computations on the encrypted data without decrypting it at any point in time during the computation (Acar, Aksu, Uluagac, & Conti 2018). In the other encryption schemes, the encrypted data needs to be decrypted first to perform the computation. The homomorphic encryption supports both additive and multiplicative homomorphism which means:

$$E(m_1+m_2) = E(m_1) + E(m_2), \text{ and } E(m_1*m_2) = E(m_1) * E(m_2)$$

where $m_1$ and $m_2$ are plain text and E is the encryption scheme. This implies that homomorphic encryption of the sum or multiplication of two numbers is equivalent to the sum or multiplication of two individually homomorphic encrypted numbers.

The homomorphic encryption scheme is mainly divided into three categories based on the number of operations that can be performed on the encrypted data:

### 3.1.1 Partially Homomorphic Encryption (PHE)

The PHE scheme allows only one type of operation either addition or multiplication an unlimited number of times on the encrypted data. Some of the examples of partially homomorphic encryption are RSA (multiplicative homomorphism) (Nisha, & Farik, 2017), ElGamal (multiplicative homomorphism) (Haraty, Otrok, & El-Kassar 2004), and Paillier (additive homomorphism) (Nassar, Erradi, & Malluhi, 2015). The PHE scheme is generally used in applications like Private Information Retrieval (PIR) and E-Voting.

### 3.1.2 Somewhat Homomorphic Encryption (SHE)

The SHE scheme allows both addition and multiplication operations but only to a limited number of times on the encrypted data. Boneh-Goh-Nissim (BGN) and Polly Cracker Scheme are some examples of the SHE scheme.

### 3.1.3 Fully Homomorphic Encryption

The FHE scheme allows all the operations like addition and multiplication an unlimited number of times on the encrypted data but it has high computational complexity and requires high-end resources for efficient implementation (Chialva, & Dooms, 2018). Gentry (Gentry, 2009) was the first one to propose the concept of FHE along with a general framework to obtain an FHE scheme. There are mainly four FHE families: Ideal lattice based, over integers (van Dijk, Gentry, Halevi, & Vaikuntanathan, 2010), Ring Learning With Errors (RLWE) based (Brakerski, & Vaikuntanathan, 2011), and NTRU-like (López-Alt, Tromer, & Vaikuntanathan 2012). We implemented Cheon-Kim-Kim-Song (CKKS) Scheme whose security is based on the hardness assumption of the RLWE problem.

## 3.2 CKKS Scheme

Cheon-Kim-Kim-Song (CKKS) is a leveled homomorphic encryption scheme that mainly works on an approximation of arithmetic numbers. It is known as leveled homomorphic encryption because there is a limit on the number of multiplications that can be performed on the encrypted data based on the selection of the parameters. It works only on the vector of real numbers but not on the scalar numbers. This scheme is based on the library Homomorphic Encryption for Arithmetic of Approximate Numbers (HEAAN) which was first introduced in (Cheon, Kim, Kim, and Song, 2017). HEAAN is an open-source homomorphic encryption library where the algorithms are implemented in C++. We used the CKKS scheme as we can encrypt the real numbers and perform the arithmetic results and get approximate or close values to the original result.

### 3.2.1 Encryption in CKKS

The encryption process happens in two steps in CKKS Scheme. In the first operation, the vector of real numbers is encoded into a plain-text polynomial. This plain text polynomial is then encrypted into a ciphertext.

### 3.2.2 Decryption in CKKS:

Similar to the encryption process, the decryption also happens in two steps. In the first operation, the ciphertext is decoded into a plain-text polynomial. This plain text polynomial is then decrypted to a vector of real numbers. Figure 1 depicts the encryption and decryption process in the CKKS scheme.

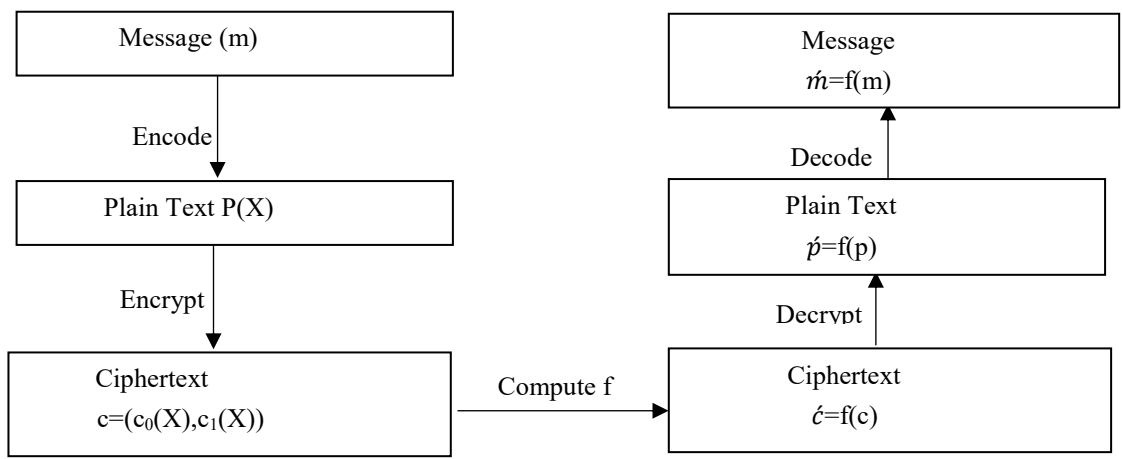

Fig 1. Block Diagram of the Encryption and Decryption in CKKS Scheme.

### 3.2.3 Parameters in CKKS:

The parameters in CKKS decide the privacy level and computational complexity of the model. These are as follows:

1. **Scaling Factor:** This defines the encoding precision for the binary representation of the number.

2. **Polynomial modulus degree:** This parameter is responsible for the number of coefficients in plain text polynomials, size of ciphertext, computational complexity, and security level. The degree should always be in the power of 2, for eg., 1024, 2048, 4096,………..
3. **Coefficient Modulus sizes:** This parameter is a list of binary sizes. A list of binary sizes of those schemes will be generated which is called coefficient modulus size. The length of the list indicates the number of multiplications possible.

### 3.2.4 Keys in CKKS

The scheme generates different types of keys which are handled by a single object called context. The keys are as follows:

1. **Secret Key:** This key is used for decryption and should not be shared with anyone.
2. **Public Encryption Key:** This key is used for the encryption of the data.
3. **Reliniearization Keys:** In general the size of the new ciphertext is 2. If there are two ciphertexts with sizes X and Y, then the multiplication of these two will result in the size getting as big as X + Y – 1. The increase in the size increases noise and also reduces the speed of multiplication. Therefore, Relinearization reduces the size of the ciphertexts back to 2 and this is done by different public keys which are created by the secret key owner.

### 3.3 Overview of the original unencrypted Wavelet Neural Networks

The WNN (Zhang, & Benveniste, 1992) has a simple architecture with just three layers, namely the input layer, hidden layer, and output layer. The input layer consists of the feature values or the explanatory variables that are introduced to the WNN and the hidden layer consists of hidden nodes which are generally referred to as Wavelons. These wavelons transform the input values into translated and dilated forms of the Mother Wavelet. The approximate target values are estimated in the output layer. All the nodes in each layer are fully connected with the nodes in the next layer. We implemented the WNN with Gaussian wavelet function as an activation function, which is defined as follows

$$f(t) = e^{-t^2} \qquad (1)$$

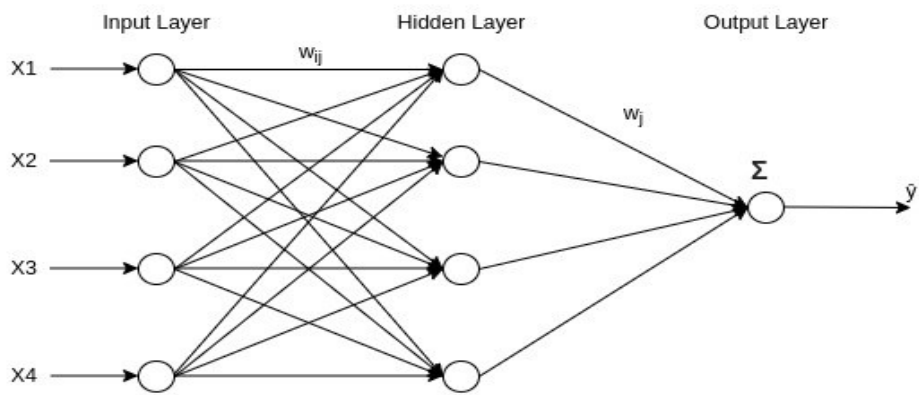

**Fig 2. Topology of a Wavelet Neural Network**

The algorithm to train the WNN is as follows. It is simpler than the backpropagation algorithm because here only the gradient descent is applied to update the parameters without backpropagating the errors (Kumar, Ravi, Mahil, & Kiran, 2008):

1. Select the number of hidden nodes and initialize all the weights, translation and dilation parameters, randomly using uniform distribution in (0,1).
2. The output value $\hat{y}$ of each sample is predicted as follows:

$$\hat{y} = \sum_{j=1}^{nhn} W_j f\left(\frac{\sum_{i=1}^{nin} w_{ij} x_{ki} - b_j}{a_j}\right) \qquad (2)$$

where *nhn* and *nin* are the numbers of hidden and input nodes respectively, $W_j$ and $w_{ij}$ are the weights between hidden to output nodes and the weights between the input to hidden nodes respectively, $b_j$ and $a_j$ are the translation and dilation parameters respectively.

3. Update the weights ($W_j$ and $w_{ij}$), translation ($b_j$), and dilation ($a_j$) parameters. The parameters of a WNN are updated by using the following formulas:

$$\Delta W_j(t+1) = -\eta \frac{\partial E}{\partial W_j(t)} + \alpha \Delta W_j(t) \qquad (3)$$

$$\Delta w_{ij}(t+1) = -\eta \frac{\partial E}{\partial w_{ij}(t)} + \alpha \Delta w_{ij}(t) \qquad (4)$$

$$\Delta a_j(t+1) = -\eta \frac{\partial E}{\partial a_j(t)} + \alpha \Delta a_j(t) \qquad (5)$$

$$\Delta b_j(t+1) = -\eta \frac{\partial E}{\partial b_j(t)} + \alpha \Delta b_j(t) \qquad (6)$$

Here the error function E is taken as Mean Squared Error (MSE),

$$E = \frac{1}{N} \sum_{i=1}^{N} (y_i - \hat{y}_i)^2 \qquad (7)$$

Where y is the actual output value, N is the number of training samples, η and α are the learning rate and momentum rate respectively.

4. The steps 2 and 3 are repeated until the error E reaches the specified convergence criteria.

## 3.4 Proposed Privacy-Preserving Wavelet Neural Network

In this paper, we proposed a novel Privacy-Preserving architecture in the form of a fully homomorphic encrypted wavelet neural network. We implemented FHE by using a library called TenSEAL [https://github.com/OpenMined/TenSEAL]. It provides a python API, but also maintains efficiency as most of its operations are implemented in C++. It performs encryption and decryption on the vector of real numbers using the CKKS scheme. It can perform various operations like addition, subtraction, multiplication, and dot product on encrypted vectors.

In our architecture, we maintained the same number of hidden nodes as the number of input nodes because the model complexity increases with the increase in the number of hidden nodes. We can also decrease the number of hidden nodes but it will lead to a decrease in the model performance and if we increase the hidden nodes, the model might perform better but the computational as well as time complexity increases.

The activation function works properly on unencrypted data but as we want to work with the encrypted data, the implementation of the exponential activation function is computationally expensive. For this reason, we performed an approximation of the activation function using Taylor series expansion which resulted in the following:

$$f(t) = 1 - t^2 + 0.5t^4 \tag{8}$$

$$\text{Where } t = \frac{\sum_{i=1}^{nin} w_{ij} x_{ki} - b_j}{a_j}$$

Accordingly, eq. 2 also gets approximated. In this architecture, the weights between the input to hidden nodes $w_{ij}$, weights between hidden to output node $W_j$, the translation parameter $b_j$, and the dilation parameter $a_j$ are all encrypted along with the input data.

The training and test phases are carried out on the encrypted data and only the parameters are decrypted and encrypted after every update to reduce the computational complexity of the model. When we pass the entire encrypted training set to the encrypted model, the computational time increases with the increase in the number of samples, and the model would take a lot of time for training. So to overcome this problem we used an optimization technique called mini-batch Stochastic Gradient Descent (SGD) (Qian, & Klabjan, 2020).

The mini-batch SGD divides the encrypted training data into a specified mini-batch size of random samples in every epoch. Our encrypted model will be trained on these mini-batches instead of the whole encrypted training set and the parameters will be updated after every mini-batch. Thus by using the mini-batch SGD the time taken by the model can be immensely reduced. The training will be stopped once the model reaches the convergence criteria which is when the change in the error is small enough to a range of 0.0001 or when the model reaches the maximum accuracy. Maximum accuracy is the highest accuracy achieved by the unencrypted model. The below block diagram explains the process flow the proposed privacy preserving WNN.

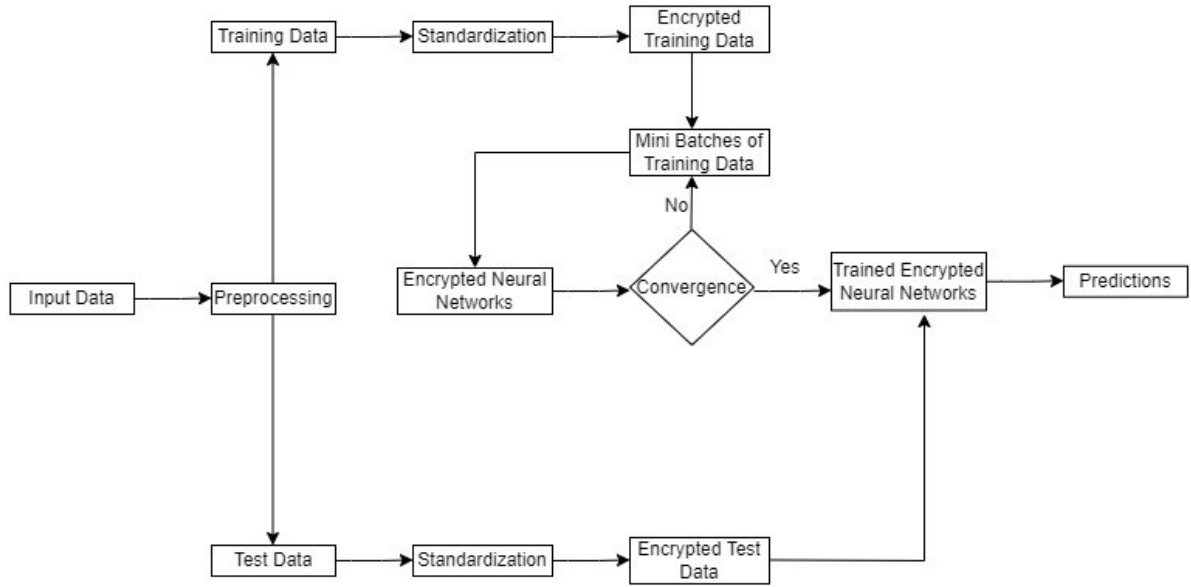

**Fig 3. Block Diagram of the Proposed PPWNN**

The algorithms 1 and 2 explain the Training and Testing procedure of the Encrypted WNN.

***Algorithm 1*** *Training the Encrypted WNN*
***Input:***
Encrypted Training Data: $E(X)$ where $X = (X_1, X_2, X_3,--------------X_n)$
Weights:- $w_{ij}$, $W_j$,
Translation and Dilation Parameters:- $b_j$, $a_j$,
Batch_size, Learning Rate ($\eta$) and Momentum ($\alpha$)
***Output:***
Encrypted Predictions:- $E(\hat{y})$, Updated Encrypted Weights = $E(w_{ij})$, $E(W_j)$
Updated Encrypted Translation and Dilation Parameters:- $E(b_j)$, $E(a_j)$
***Function*** TrainEncryptedWNN($E(X)$, $w_{ij}$, $W_j$, $b_j$, $a_j$)
    1. **While** true
    2.     **if** $\Delta E >= 0.0001$ or accuracy <= max_accuracy
    3.         Encrypt the parameters $w_{ij}$, $W_j$, $b_j$, $a_j$
            Divide $E(X)$ into random batches of specified Batch_size
    4.         **for** each sample in Batch_size
    5.             Generate $E(\hat{y})$ with (2)
    6.             Calculate the MSE by using (7) and derivatives for each sample
    7.             Add all derivatives
                Update the parameters $E(w_{ij})$, $E(W_j)$, $E(b_j)$, $E(a_j)$ with (3), (4), (5), (6)
    8.         Decrypt the parameters $E(w_{ij})$, $E(W_j)$, $E(b_j)$, $E(a_j)$
    9.     **else**
    10.        Break

In the above algorithm, ΔE is the change in the Mean Squared Error of the current batch and previous batch and max_accuracy is the maximum accuracy obtained by the unencrypted model

```
Algorithm 2 Testing the Encrypted WNN
Input:
Encrypted Test Data: E(X`) where X = (X₁`, X₂`, X₃`,--------------Xₙ`)
Updated Encrypted Weights: E(wᵢⱼ`), E (Wⱼ`)
Updated Encrypted Translation and Dilation Parameters: E(bⱼ`), E(aⱼ`)
Output:
Encrypted Predictions: E(ŷ)
Function Test EncryptedWNN(E(X`), E(wᵢⱼ`), E(Wⱼ`), E(bⱼ`), E(aⱼ`))
    1. for each sample in Batch_size
    2.     Generate E(ŷ) with (2)
    3. Decrypt the predictions and calculate the accuracy and AUC.
```

## 4 Datasets Description

The features of all the datasets are presented in the Appendix.

### 4.1 Health Care Datasets

#### 4.1.1 Haberman's Survival Dataset

The dataset contains samples collected from a study which was conducted between the years 1958 and 1970 at the University of Chicago's Billing Hospital on the patients who had survived after undergoing surgery for breast cancer. This dataset has 306 instances and 4 features which include the target variable, namely, Survival Status of the Patient (Haberman, 1976). Out of 306 instances, 225 instances are the patients who survived more than 5 years and 81 instances are the people who died within 5 years. The description of features is provided in Table A.1.

#### 4.1.2 Breast Cancer Coimbra Dataset

In this dataset, the features are anthropometric data and parameters which are generally gathered in a routine blood analysis. There are a total of 10 features including the target variable, namely, presence or absence of breast cancer and 116 instances in this dataset (Patrício, Pereira, & Crisóstomo, et al., 2018). Out of the 116 instances, 52 instances are the people who are healthy and 64 instances are the people who are the risk of breast cancer. The description of the features is provided in Table A.2.

#### 4.1.3 Fertility Dataset

This dataset contains information about the semen samples provided by 100 volunteers. This dataset has 100 instances and 10 features which includes the target variable, namely, whether the diagnosis was Normal or Altered (Méndez, 2012). Out of the 100 instances, 88 instances are the samples whose output was Normal and 12 instances are the samples whose output was Altered. The description of the features is provided in Table A.3.

### 4.1.4 Heart Disease Dataset

This dataset has 303 instances and 14 features which includes the target variable, namely, whether the person has heart disease or not. Out of the 303 instances, 138 instances are the people who would not be affected by heart disease and 165 instances are the people who would be affected by heart disease. The description of the features is provided in Table A.4.

### 4.1.5 Diabetes Dataset

This dataset is mainly for the female gender and has 768 instances and 9 features that include the target variable, namely, whether the patient is diabetic or not (Smith, et.al., 1988). Out of the 768 instances, 500 instances are the people who are negative for diabetes, and 268 are the instances who are positive for diabetes. The description of the features is provided in Table A.5.

## 4.2 Financial Datasets

### 4.2.1 BankNote Authentication Dataset

In this dataset, the data were extracted from images that were taken from genuine and forged banknotes. Wavelet transform was used to extract features from images. This dataset has 1372 instances and 5 features including the target variable, namely, whether the Note is authentic or not (Dua, & Graff, 2019). Out of 1372 instances, 762 instances are the Notes that are genuine and 610 instances are the Notes which are forged. The description of the features is provided in Table A.6.

### 4.2.2 Qualitative Bankruptcy Dataset

In this dataset, there are 250 instances and 7 features including the target variable, namely, whether, a bank is bankrupt or non-bankrupt (Kim, & Ingoo 2003). Out of the 250 instances, 143 instances are non-bankrupt banks and 107 are bankrupt. The description of the features is provided in Table A.7.

# 5 Results and discussion

All the experiments are carried out on a system with the following configuration: HP Z8 workstation with Intel Xeon (R) Gold 6235R CPU processor, Ubuntu 20.04lts, and having RAM of 376.6 GB. The number of hidden nodes is kept the same as the number of input nodes. Accuracy and Area Under the Receiver Operating Characteristics Curve (AUC) are taken as the performance metrics. Standardization was applied to the features in the following datasets: Haberman's Survival, BankNote Authentication, Breast Cancer, Heart Disease, and Diabetes Prediction Dataset. The fertility dataset had a class imbalance with a class distribution of Class '0': 88 versus Class '1': 12. To balance the dataset we applied the data balancing technique called SMOTE (Chawla, Bowyer, Hall, & Kegelmeyer, 2002) to make the data perfectly balanced with 88 samples in each class. Standardization was performed on the features and all the features were considered for training the model. In the Qualitative Bankruptcy dataset, all the features are categorical. We converted the labels of all the features into a numeric form.

## 5.1 Health Care Dataset Results:

In the health care datasets, the results of the Fertility Dataset and Diabetes Prediction Dataset in both the Unencrypted and Encrypted Wavelet Neural Network yielded nearly equal performance as the Accuracy and AUC are almost same in both the cases. In the other datasets, the encrypted model has performed better compared to the unencrypted model. This might be due to the approximation of the activation function and also the convergence criteria might be one of the factors.

## 5.2 Financial Dataset Results:

In the financial datasets, both the unencrypted and encrypted models performed similarly as the Accuracy and AUC are nearly equal.

**Table.1 Health Care Dataset Results**

| Datasets | Unencrypted Mini Batch SGD WNN | | Encrypted Mini Batch SGD WNN | | The average time taken for each epoch in the Encrypted Mini Batch SGD WNN |
|---|---|---|---|---|---|
| | Accuracy | AUC | Accuracy | AUC | |
| Haberman's Survival | 0.48 | 0.44 | 0.54 | 0.55 | 1 mn 23 secs |
| Fertility Dataset | **0.63** | **0.63** | **0.61** | **0.62** | **9 mins 3 secs** |
| Heart disease Dataset | 0.54 | 0.50 | 0.75 | 0.74 | 16 mins 20 secs |
| Breast Cancer Dataset | 0.46 | 0.5 | 0.62 | 0.61 | 6 mins 16 secs |
| Diabetes Prediction Dataset | **0.52** | **0.53** | **0.52** | **0.50** | **8 mins 41 secs** |

**Table.2 Financial Dataset Results**

| Datasets | Unencrypted Mini Batch SGD WNN | | Encrypted Mini Batch SGD WNN | | The average time taken for each epoch in the Encrypted Mini Batch SGD WNN |
|---|---|---|---|---|---|
| | Accuracy | AUC | Accuracy | AUC | |
| BankNote Authentication | **0.49** | **0.50** | **0.50** | **0.49** | **5 mins 1 sec** |
| Qualitative Bankruptcy | **0.58** | **0.50** | **0.54** | **0.55** | **3 mins 25 sec** |

# 6 Conclusions

A fully Homomorphic Encrypted Wavelet Neural Network is proposed with SGD and an approximated activation function. The model provides high security because along with input data, the predictions, the parameters like weights, translation, and dilation parameters are also encrypted. But there is a limitation to the model in that the average time per epoch increases with the increase in the number of samples and features. Moreover, homomorphically encrypted networks require a high computational power system for efficient execution. Finally, we can conclude that the data is well protected throughout the process but it comes at the cost of resources and time. The idea of a privacy-preserving WNN can be extended in a Federated Learning setup where the model used on the individual nodes can be our PPWNN which will ensure that the privacy of the data is protected even on the individual machines. Further, this PPWNN can also be employed to solve regression problems.

# Appendix A:

Table A.1 Haberman's Survival Dataset

| | Feature Description |
|---|---|
| 1. | Age of patient at the time of Operation |
| 2. | Patient's year of Operation |
| 3. | Number of Positive Axillary Nodes Detected |
| 4. | Survival Status |

Table A.2 Breast Cancer Dataset

| | Feature Description |
|---|---|
| 1. | Age of the Patient |
| 2. | Body Mass Index (kg/m$^2$) |
| 3. | The glucose level in the body (mg/dL) |
| 4. | Insulin level in the body (μU/mL) |
| 5. | Homeostasis Model Assessment |
| 6. | Leptin (ng/mL) |
| 7. | Adiponectin (μg/mL) |
| 8. | Resistin (ng/mL) |
| 9. | Chemokine Monocyte Chemoattractant Protein 1 (MCP-1) |
| 10. | Classification |

Table A.3 Breast Cancer Dataset

| | Feature Description |
|---|---|
| 1. | Season in which the Analysis was performed |
| 2. | Age at the time of analysis (18-36) |
| 3. | Childish Disease (chicken pox, measles, mumps, polio) |
| 4. | Accident or Serious Trauma |
| 5. | Surgical Intervention |
| 6. | High Fevers in the last year |
| 7. | Frequency of Alcohol Consumption |
| 8. | Smoking Habit |
| 9. | Number of Hours spent sitting per day |
| 10. | Output |

Table A.4 Heart Disease Dataset

| # | Feature Description |
|---|---|
| 1. | Age of the Person |
| 2. | Sex (Male or Female) |
| 3. | Chest Pain Type |
| 4. | Trestbps: resting blood pressure (in mm Hg on admission to the hospital) |
| 5. | Cholesterol |
| 6. | Fbs (fasting blood sugar > 120 mg/dl) |
| 7. | restecg: resting electrocardiographic results |
| 8. | thalach (maximum heart rate achieved) |
| 9. | exang (exercise induced angina) |
| 10. | Oldpeak (ST depression induced by exercise relative to rest) |
| 11. | Slope (The slope of the peak ecercise ST segment) |
| 12. | ca (number of major vessels covered by fluoroscopy) |
| 13. | thal |
| 14. | target (diagnosis of heart disease) |

Table A.5 Diabetes Dataset

| # | Feature Description |
|---|---|
| 1. | Pregnancies (Number of times pregnant) |
| 2. | Glucose (Oral Glucose Tolerance Test result) |
| 3. | Blood Pressure (Diastolic Blood Pressure values in (mm Hg)) |
| 4. | SkinThickness (Triceps skin fold thickness in (mm)) |
| 5. | Insulin (2-Hour serum Insulin (μU/ml) |
| 6. | BMI (Body Mass Index) |
| 7. | Diabetes Pedigree Function |
| 8. | Age (Age in Years) |
| 9. | Class |

Table A.6 BankNote Authentication Dataset

| # | Feature Description |
|---|---|
| 1. | The variance of Wavelet Transformed Image |
| 2. | The skewness of Wavelet Transformed Image |
| 3. | Curtosis of Wavelet Transformed Image |
| 4. | Entropy of Image |
| 5. | Class |

Table A.7 Qualitative Bankruptcy Dataset

| # | Feature Description |
|---|---|
| 1. | Industrial Risk |
| 2. | Management Risk |
| 3. | Financial Flexibility |
| 4. | Credibility |
| 5. | Competitiveness |
| 6. | Operating Risk |
| 7. | Class |